\documentclass[11pt,a4paper]{article}
\usepackage[hyperref]{emnlp2018}
\usepackage{times}
\usepackage{latexsym}

\usepackage{url}

\aclfinalcopy 

\usepackage{amssymb}
\usepackage{amsmath}
\usepackage{graphicx}

\usepackage{bm}
\newcommand{\vect}[1]{\bm{#1}}
\newcommand{\matr}[1]{\bm{#1}}

\newcommand{\va}[0]{\vect{a}}
\newcommand{\vb}[0]{\vect{b}}
\newcommand{\vc}[0]{\vect{c}}

\newcommand{\vh}[0]{\vect{h}}
\newcommand{\vv}[0]{\vect{v}}
\newcommand{\vx}[0]{\vect{x}}
\newcommand{\vz}[0]{\vect{z}}

\newcommand{\vs}[0]{\vect{s}}

\newcommand{\vy}[0]{\vect{y}}
\newcommand{\vg}[0]{\vect{g}}

\newcommand{\vr}[0]{\vect{r}}
\newcommand{\vp}[0]{\vect{p}}
\newcommand{\vtheta}[0]{\vect{\theta}}

\newcommand{\mW}[0]{\matr{W}}

\newcommand{\mE}[0]{\matr{E}}
\newcommand{\mH}[0]{\matr{H}}

\newcommand{\mX}[0]{\matr{X}}

\newcommand{\mY}{\matr{Y}}
\usepackage{color}
\definecolor{olive}{RGB}{0,184,136}

\title{UMONS Submission for WMT18 Multimodal Translation Task}

 \author{Jean-Benoit Delbrouck \and St\'ephane Dupont \\
         TCTS Lab, University of Mons, Belgium\\
          \{jean-benoit.delbrouck, stephane.dupont\}@umons.ac.be}

\date{}

\begin{document}
\maketitle
\begin{abstract}
This paper describes the UMONS solution for the Multimodal Machine Translation Task presented at the third conference on machine translation (WMT18). We explore a novel architecture, called deepGRU, based on recent findings in the related task of Neural Image Captioning (NIC). The models presented in the following sections lead to the best METEOR translation score for both constrained (English, image) $\rightarrow$ German and (English, image) $\rightarrow$ French sub-tasks.
\end{abstract}

\section{Introduction}

	In the field of Machine Translation (MT), the efficient integration of multimodal information still remains a challenging task. It requires combining diverse modality vector representations with each other. These vector representations, also called context vectors, are computed in order the capture the most relevant information in a modality to output the best translation of a sentence.  \\
	
	To investigate the effectiveness of information obtained from images, a multimodal neural machine translation (MNMT) shared task \cite{specia-EtAl:2016:WMT} has been  introduced to the community.\footnote{\url{http://www.statmt.org/wmt18/multimodal-task.html}} Even though soft attention models had been extensively studied in MNMT \cite{delbrouck-dupont:2017:EMNLP2017, caglayan2016does, Calixto-1175}, the most successful recent work \citep{W17-4746_ozan} focused on using the max-pooled features extracted from a convolutional network to modulate some components of the system (i.e. the target embeddings). Convolutional features or attention maps recently showed some success \cite{DBLP:journals/corr/abs-1712-03449} in a encoder-based attention model conditioned on the source encoder representation. Both model types lead to similar results, the latter being slightly complex and taking longer to train. One similar feature they share is that the proposed models remain relatively small. Indeed, the number of trainable parameters seems ``upper bounded" due to the number of unique training examples being limited (cfr. Section \ref{data}). Heavy or complex attention models on visual features showed premature convergence  and restricted scalability. \\
	
	The model proposed by the University of Mons (UMONS) in 2018 is called DeepGRU, a novel idea based on the previously investigated conditional GRU (cGRU).\footnote{\url{https://github.com/nyu-dl/dl4mt-tutorial/blob/master/docs/cgru.pdf}} We enrich the architecture with three ideas borrowed from the closely related NIC task: a third GRU as bottleneck function, a multimodal projection and the use of gated tanh activation. We make sure to keep the overall model light, efficient and rapid to train. We start by describing the baseline model in Section \ref{baseline} followed by the three aforementioned NIC upgrades which make up our
  deepGRU model in Section \ref{multimodgated}. Finally, we present the data made available by the Multimodal Machine Translation Task in Section \ref{data} and the results in section \ref{results}, then engage a quick discussion in Section \ref{concl}.
	
\section{Baseline Architecture} \label{baseline}

Given a source sentence $\mX = (\vx_1, \vx_2, \hdots , \vx_M)$ and an image $I$, an attention-based encoder-decoder model \cite{BahdanauCB14} outputs the translated sentence $\mY = (\vy_1, \vy_2, \hdots , \vy_N)$. If we denote $\vtheta$ as the model parameters, then $\vtheta$ is learned by maximizing the likelihood of the observed sequence $\mY$ or in other words by minimizing the cross entropy loss. The objective function is given by:
\begin{equation}
\mathcal{L}(\vtheta) = - \sum\limits_{t=1}^n \log p_{\vtheta}(\vy_t | \vy_{<t}, I, X) \label{eq:5}
\end{equation}

Three main components are involved: an encoder, a decoder and an attention model. \\

\textbf{Encoder} \quad  At every time-step $t$, an encoder creates an annotation $\vh_t$ according to the current embedded word $\vx_t$ and internal state $\vh_{t-1}$:

\begin{equation}
    \vh_t = f_{\text{enc}}(\vx_t^{\prime}, \vh_{t-1})
\end{equation}

Every word $\vx_t$ of the source sequence $\mX$ is an index in the embedding matrix $\mE^x$ so that the following formula maps the word to the $f_{\text{enc}}$ size $S$:

\begin{equation}
\vx_t^{\prime} = \mW^x \mE^x x_{t}
\end{equation}
The total size of the embeddings matrix $\mE^x$ depends on the source vocabulary size $|\mathcal{Y}_s|$ and the embedding dimension $d$ such that $\mE^x \in \mathbb{R}^{|\mathcal{Y}_s|\times d}$. The mapping matrix $\mW^x$ also depends on the embedding dimension  because $\mW^x \in \mathbb{R}^{d \times S}$. \\

The encoder function $f_{\text{enc}}$ is a bi-directional GRU \cite{cho-al-emnlp14}. The following equations define a single GRU block (called $f_{\text{gru}}$ for future references) :

\begin{align} \label{grucell}
\vz_t =& ~ \sigma \left( \vx_t^{\prime} + \mW^z \vh_{t-1} \right) \nonumber \\
\vr_t =& ~ \sigma \left( \vx_t^{\prime} + \mW^r \vh_{t-1} \right) \nonumber \\
\underline{\vh}_t =& ~ \text{tanh} \left(  \vx_t^{\prime}  + \vr_t \odot (\mW^h \vh_{t-1} )  \right)  \nonumber \\       
\vh_t^{\prime} =& ~ (1 - \vz_t) \odot \underline{\vh}_t + \vz_t \odot \vh_{t-1}
\end{align}	
where $\vh_t^{\prime} \in \mathbb{R}^S$.
Our encoder consists of two GRUs, one is reading the source sentence from 1 to M and the second from M to 1. The final encoder annotation $\vh_t$ for timestep $t$ becomes the concatenation of both GRUs annotations $\vh_t^{\prime}$. Therefore, the encoder set of annotations $\mH$ is of size $M \times 2S$. \\

\textbf{Decoder} \quad At every time-step $t$, a decoder outputs probabilities $\vp_t$ over the target vocabulary $\mathcal{Y}_d$ according to previously generated word $\vy_{t-1}$, internal state $\vs_{t-1}$ and image $I$:

\begin{equation}
y_t \sim \vp_t = f_{\text{bot}}\big(\hspace{0.5em} f_{\text{dec}}(\vy_{t-1}, \vs_{t-1}, I) \hspace{0.5em} \big)
\end{equation} \\

Every word $\vy_t$ of the target sequence $\mY$ is an index in the embedding matrix $\mE^y$ so that the following formula maps the word in the $f_{\text{dec}}$ size $D$:

\begin{equation}
\vy_t^{\prime} = \mW^y \mE^y \vy_{t-1}
\end{equation}

The decoder function $f_{\text{dec}}$ is a conditional GRU (cGRU). The following equations describes a cGRU cell :

\begin{align} \label{cgrucell}
\vs_t^{\prime} =& ~ f_{\text{gru}_1}(\vy_t^{\prime},\vs_{t-1}) \nonumber \\
\vc_t =& ~ f_{\text{att}}(\vs_t^{\prime},I, \mH) \nonumber \\
\vs_t =& ~ f_{\text{gru}_2}(\vs_t^{\prime},\vc_t) 
\end{align}	

where $f_{\text{att}}$ is the visual attention module over the set of source annotation $\mH$ and pooled vector $\vv$ of ResNet-50 features extracted from image $I$. More precisely, our attention model is the product between the so-called soft attention over the $M$ source annotations $\vh_{\{0, \hdots, M-1\}}$ and the linear transformation over pooled vector $v$ of image I :

\begin{align}
\va_t^{\prime}  =& \mW^a  \tanh(\mW^{\text{s}} \vs_t^{\prime} + \mW^{\text{H}} \mH) \\
\va_t =& \text{softmax} ( \va_t^{\prime} )  \\
\vc_t^{\prime} =& ~ \sum_{i=0}^{M-1} \va_{t_i} \vh_i \\
\vv_t =& ~  \tanh ( \mW^{\text{img}} I  ) \label{eq:vt} \\
\vc_t =& ~ \mW^c \vc_t^{\prime} \odot \vv_t   \label{eq:vt_merge}
\end{align}

The bottleneck function $f_{\text{bot}}$ projects the cGRU output into probabilities over the target vocabulary. It is defined so:

\begin{align}
\vb_t &= \tanh(\mW^{\text{bot}}[\vy_{t-1}, \vs_{t},\vc_t] \label{f_bt})  \\ 
y_t \sim \vp_t &= \text{softmax}(\mW^{\text{proj}} \vb_t) \label{f_proj}
\end{align}

where $[\cdot, \cdot]$  denotes the concatenation operation. \\

\section{DeepGRU} \label{multimodgated}

The deepGRU decoder \cite{jbcaption} is a variant of the cGRU decoder. \\

\textbf{Gated hyperbolic tangent} \quad First, we make use of the gated hyperbolic tangent activation \cite{Teney2017TipsAT} instead of tanh. 
This non-linear layer implements a function $f_{\text{ght}} : x \in \mathbb{R}^n \rightarrow y \in \mathbb{R}^m$ with parameters defined as follows:

\begin{align} \label{ght}
\vy^{\prime} =& ~ \tanh(\mW^t\vx+b) \nonumber \\
\vg =& ~ \sigma(\mW^g\vx+b) \nonumber \\
\vy =& ~ \vy^{\prime} \odot  \vg
\end{align}	

where $\mW^x,\mW^g \in \mathbb{R}^{n\times m}$. We apply this gating system for equation \ref{eq:vt} and \ref{f_bt}. \\

\textbf{GRU bottleneck} \quad When working with small dimensions, one can afford to replace the computation of $\vb_t$ of equation \ref{f_bt} by a new gru block $f_{\text{gru}}$:

\begin{equation}
\vb_t^{v} = f_{ght} \big(\mW_{\text{bot}}^v \big( f_{\text{gru}_3}([\vy_{t-1}, \vs_{t}^{\prime},\vv_t], \vs_t) \big) \big) \label{eq:bottgru}
\end{equation}

The GRU bottleneck can be seen as a new block $f_{\text{gru}_3}$ encoding the visual information $\vv_t$ with its surrounding context ($\vy_{t-1}$ and $\vs_{t}^{\prime}$). Therefore, equation $\ref{eq:vt_merge}$ is not computed with $\vv_t$ anymore so that the second block $f_{\text{gru}_2}$ encodes the textual information only. \\

\textbf{Multimodal projection} \quad Because we now have a linguistic GRU block and a visual GRU block, we want both representations to have their own projection to compute the candidate probabilities. Equation \ref{f_bt} and \ref{f_proj} becomes:

\begin{align}
\vb_t^t &= f_{ght} (\mW_{\text{bot}}^t \; \vs_t) \label{eq:bott_st} \\
y_t \sim \vp_t &= \text{softmax}(\mW^t_{\text{proj}} \; \vb_t^t + \mW^v_{\text{proj}} \; \vb_t^{v})
\end{align}

where $\vb_t^{v}$ comes from equation \ref{eq:bottgru}. Note that we use the gated hyperbolic tangent for equation \ref{eq:bottgru} and \ref{eq:bott_st}. \\

\section{Data and settings} \label{data}

The Multi30K dataset \citep{elliott-EtAl:2016:VL16} is provided by the challenge. For each image, one of the English descriptions was selected and manually translated into German and French by a professional translator. As training and development data, 29,000 and 1,014 triples are used respectively. We use the three available test sets to score our models. The Flickr Test2016 and the Flickr Test2017 set contain 1000 image-caption pairs and the ambiguous MSCOCO test set \citep{ElliottFrankBarraultBougaresSpecia2017} 461 pairs. For the WMT18 challenge, a new Flickr Test2018 set of 1,071 sentences is released without the German and French gold translations. \\
	
Marices of the model are initialized using the Xavier method \cite{Glorot10understandingthe} and the gradient norm is clipped to 5. We chose ADAM \citep{kingma2014adam} as the optimizer with a learning rate of 0.0004 and batch-size 32. To marginally reduce our vocabulary size, we use the byte pair encoding (BPE) algorithm on the train set to convert space-separated tokens into sub-words \cite{P16-1162}. With 10K merge operations, the resulting vocabulary sizes of each language pair are: 5204 $\rightarrow$ 7067 tokens for English$\rightarrow$ German and 5835$\rightarrow$ 6577 tokens for English$\rightarrow$French.\\

We use the following regularization methods: we apply dropout of 0.3 on source embeddings $\vx^{\prime}$, 0.5 on source annotations $\mH$ and 0.5 on both bottlenecks $\vb_t^t$ and $\vb_t^v$. We also stop training when the METEOR score does not improve for 10 evaluations on the validation set (i.e.~one validation is performed every 1000 model updates). \\

The dimensionality of the various settings and layers is as follows: \\ Embedding size $d$ is 128, encoder and decoder GRU size $S$ is 256, embedding layers are: $[\mW^x,\mW^y \in \mathbb{R}^{128 \times 256}$, $\mH = M \times 512$, $\mE^x \in \mathbb{R}^{\mathcal{Y}_s \times 128},  \mE^y \in \mathbb{R}^{\mathcal{Y}_d \times 128}]$. \\

Attention matrices: $[\mW^s \in \mathbb{R}^{256 \times 512}, \mW^H \in \mathbb{R}^{512 \times 512}, \mW^a \in \mathbb{R}^{512 \times 1}, \mW^c \in \mathbb{R}^{512 \times 256}, \mW^{\text{img}} \in \mathbb{R}^{2048 \times 256}]$. \\

Bottleneck matrices: $[\mW_{\text{bot}}^t, \mW_{\text{bot}}^v  \in \mathbb{R}^{256 \times 128}]$ and projection matrices: $[\mW^t_{\text{proj}}, \mW^v_{\text{proj}} \in \mathbb{R}^{128 \times \mathcal{Y}_d}]$. Weights $\mE^y$ and $\mW^t_{\text{proj}}$ are tied.\\

The size of gated hyperbolic tangent weights $\mW^t, \mW^g$ depends on their respective application.
\section{Results} \label{results}

Our models performance are evaluated according to the following automated metrics: BLEU-4 \cite{Papineni:2002:BMA:1073083.1073135} and METEOR \cite{denkowski:lavie:meteor-wmt:2014}. We decode with a beam-search of size 12 and use model ensembling of size 5 for German and 6 for French. We used the nmtpytorch \cite{nmtpy2017} framework for all our experiments. We also release our code.\footnote{\url{https://github.com/jbdel/WMT18_MNMT}} \\

\begin{tabular}{llllll}
			\multicolumn{1}{c}{\bf Test sets}  &\multicolumn{1}{l}{\bf 	BLEU}   &\multicolumn{1}{l}{\bf 	METEOR} 
			\\ \hline \\
            \textbf{Test 2016 Flickr} \\
			\\ 
            FR-Baseline & 59.08 & 74.73  \\
            FR-DeepGRU & 62.49 \footnotesize{+3.41} &  76.83 \footnotesize{+2.10} \\
            DE-Baseline & 38.43 &   58.37 \\
            DE-DeepGRU & 40.34 \footnotesize{+1.91}&  59.58 \footnotesize{+1.21}  \\
            \\
            \textbf{Test 2017 Flickr} \\
			\\ 
            FR-Baseline & 51.86 &  72.75  \\
            FR-DeepGRU & 55.13 \footnotesize{+3.27} &  71.52 \footnotesize{+1.98}\\
            DE-Baseline & 30.80 &  52.33  \\
            DE-DeepGRU & 32.57 \footnotesize{+1.77} &  53.60 \footnotesize{+1.27} \\
            \\
           \textbf{Test 2017 COCO} \\
			\\ 
            FR-Baseline & 43.31 & 64.39  \\
            FR-DeepGRU & 46.16  \footnotesize{+2.85} &  65.79 \footnotesize{+1.40} \\
            DE-Baseline & 26.30 &  48.45  \\
            DE-DeepGRU & 29.21 \footnotesize{+2.91} &  49.45 \footnotesize{+1.00} \\
            \\
           \textbf{Test 2018 Flickr} \\
			\\ 
            FR-DeepGRU & 39.40 & 60.17 \\    
            DE-DeepGRU & 31.10 & 51.64 \\

\end{tabular}

\section{Conclusion and future work}\label{concl}

The full leaderboard scores \footnote{\url{https://competitions.codalab.org/competitions/19917}} shows close results and it seems that everybody converges towards the same translation quality score. A few questions arise. Did we reach ---to some extent--- the full potential of images related to the information they can provide? Should we try and add traditional machine translation techniques such as post-edition, since images have been exploited successfully? Another major step forward would be to successfully develop strong and stable models using convolutional features, the latter having 98 times more features than the max-pooled ones.

\bibliographystyle{acl_natbib_nourl}
\bibliography{emnlp2018}

\end{document}